\documentclass[conference]{IEEEtran}
\IEEEoverridecommandlockouts

\usepackage{cite}
\usepackage{amsmath,amssymb,amsfonts}
\usepackage{algorithmic}
\usepackage{graphicx}
\usepackage{textcomp}
\usepackage{xcolor}
\usepackage{booktabs} 
\usepackage{tabularx}
\usepackage{makecell}
\usepackage{xcolor}
\usepackage{multirow}
\newcolumntype{Y}{>{\centering\arraybackslash}X}

\def\BibTeX{{\rm B\kern-.05em{\sc i\kern-.025em b}\kern-.08em
    T\kern-.1667em\lower.7ex\hbox{E}\kern-.125emX}}
\begin{document}

\title{MetaSSP: Enhancing Semi-supervised Implicit 3D Reconstruction through Meta-adaptive EMA and SDF-aware Pseudo-label Evaluation\\}

\author{
\IEEEauthorblockN{Luoxi Zhang}
\IEEEauthorblockA{\textit{Doctoral Program in Empowerment Informatics} \\
\textit{University of Tsukuba}\\
Tsukuba, Ibaraki, Japan \\
zhang.luoxi@image.iit.tsukuba.ac.jp}
\and
\IEEEauthorblockN{Chun Xie}
\IEEEauthorblockA{\textit{Center for Computational Science} \\
\textit{University of Tsukuba}\\
Tsukuba, Ibaraki, Japan \\
xiechun@ccs.tsukuba.ac.jp}
\and
\IEEEauthorblockN{Itaru Kitahara}
\IEEEauthorblockA{\textit{Center for Computational Science} \\
\textit{University of Tsukuba}\\
Tsukuba, Ibaraki, Japan \\
kitahara@ccs.tsukuba.ac.jp}
}

\maketitle

\begin{abstract}
Implicit SDF-based methods for single-view 3D reconstruction achieve high-quality surfaces but require large labeled datasets, limiting their scalability. We propose MetaSSP, a novel semi-supervised framework that exploits abundant unlabeled images. Our approach introduces gradient-based parameter importance estimation to regularize adaptive EMA updates and an SDF-aware pseudo-label weighting mechanism combining augmentation consistency with SDF variance. Beginning with a 10\% supervised warm-up, the unified pipeline jointly refines labeled and unlabeled data. On the Pix3D benchmark, our method reduces Chamfer Distance by approximately 20.61\% and increases IoU by around 24.09\% compared to existing semi-supervised baselines, setting a new state of the art.
\end{abstract}


\begin{figure*}[t]  
    \centering  
    \includegraphics[width=\linewidth]{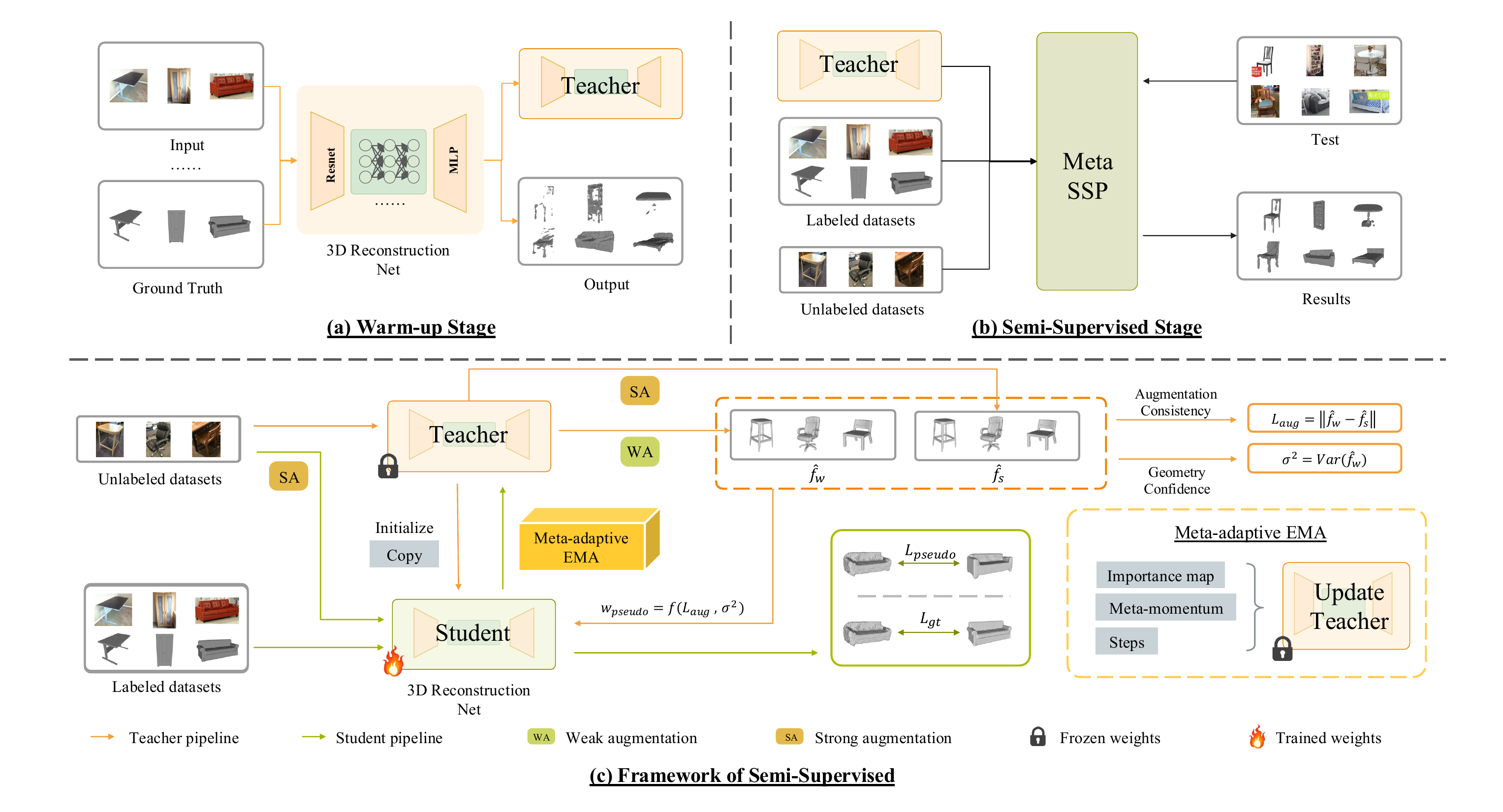}  
    \caption{MetaSSP Semi-Supervised 3D Reconstruction Pipeline: (a) Supervised Warm-up Stage; (b) Semi-Supervised Knowledge Distillation Stage; (c) Detailed Teacher–Student Meta-Adaptive EMA and Pseudo-Label Weighting Mechanism.
    }  
    \label{fig:framework}  
\end{figure*}

\section{Introduction}

Recovering high-fidelity 3D geometry and appearance from a single RGB image—known as single-view 3D reconstruction—has become a key problem in applications such as augmented reality, robotics, and virtual production. Implicit surface representations, particularly those based on Signed Distance Functions (SDF), have shown strong capability in capturing fine details and producing watertight surfaces \cite{DeepSDF, NeuS}. However, state-of-the-art implicit approaches rely heavily on large-scale labeled 3D datasets \cite{pix3d,gso,front3d}, whose creation is costly and time-consuming, thus limiting their scalability to new domains and real-world scenes \cite{DP-AMF,MGP-KAD,M3D}.

Semi-supervised learning (SSL) offers a practical solution by leveraging abundant unlabeled data alongside a small set of labeled samples. While effective in 2D and point-based 3D tasks, existing SSL methods—such as MonoDepth2 \cite{monodepth2} and PointContrast \cite{PointContrast}—struggle to generalize to implicit SDF representations. This difficulty arises from two factors: (1) conventional EMA-based teacher–student training easily becomes unstable as the parameter gap widens over time \cite{MeanTeacher}, and (2) pseudo-label reliability is hard to estimate in continuous SDF fields, where minor geometric deviations are difficult to detect \cite{FixMatch}.

To address these challenges, we propose MetaSSP, a semi-supervised framework specifically designed for SDF-based implicit reconstruction. MetaSSP introduces a meta-adaptive EMA mechanism that stabilizes teacher–student learning by combining gradient-guided parameter regularization with dynamic momentum control. Additionally, an SDF-aware pseudo-label weighting strategy fuses augmentation consistency and SDF variance to suppress noisy supervision from unreliable unlabeled data. Starting from a 10\% labeled warm-up, our unified pipeline jointly optimizes labeled and unlabeled samples. Experiments on the Pix3D benchmark show that MetaSSP reduces Chamfer Distance by 20.61\% and increases IoU by 24.09\%, achieving state-of-the-art performance under limited supervision.

Our main contributions are summarized as follows:
\begin{itemize}
\item \textbf{Meta-adaptive EMA regularization.} We introduce a teacher–student update scheme that jointly leverages gradient-based parameter importance and dynamic momentum scheduling to stabilize semi-supervised training.
\item \textbf{SDF-aware pseudo-label weighting.} We design a reliability weighting mechanism that integrates augmentation consistency and SDF variance to filter noisy pseudo-labels.
\item \textbf{Semi-supervised SDF reconstruction framework.} We present the first unified SSL pipeline tailored for implicit SDF reconstruction, achieving state-of-the-art performance on Pix3D with only 10\% labeled data.
\end{itemize}

\section{Related Work}

\subsection{Semi-supervised Learning in 3D Reconstruction}
Semi-supervised learning has been applied to single-view 3D reconstruction to reduce reliance on costly annotations. SSP3D introduced an attention-guided prototype shape prior and a discriminator to judge pseudo-labels, improving voxel-based reconstruction under limited labels \cite{SSP3D}. Zhou \emph{et al.} proposed SSMP, which fuses multiple shape priors and employs self-attention to refine point-cloud predictions in a semi-supervised setting \cite{SSMP}. However, these methods target explicit depth or point representations and do not extend to SDF-based implicit networks, where continuous geometry poses additional challenges for consistency and prior guidance.

\subsection{EMA Strategies and Parameter Importance Estimation}
Teacher–student frameworks often use Exponential Moving Average (EMA) to stabilize pseudo-label generation. The Mean Teacher model first demonstrated EMA’s effectiveness for semi-supervised classification \cite{MeanTeacher}. More recently, Reflective Teacher applied a regularizer to prevent catastrophic forgetting during EMA updates in 3D object detection \cite{ReflectiveTeacher}. Despite these advances, existing EMA strategies lack per-parameter adaptation or dynamic momentum control \cite{NoisyStudent}, limiting their stability when teacher and student diverge in SDF-based reconstruction.

\subsection{Pseudo-labeling and Its Limitations in 3D Reconstruction}
Pseudo-labeling methods assign labels to unlabeled data based on model confidence or auxiliary modules. FixMatch demonstrated strong-augmented consistency for 2D classification, using high-confidence predictions as pseudo-labels \cite{FixMatch}, ShapeHD \cite{ShapeHD} employs an adversarial discriminator for 3D shape realism. In 3D reconstruction, SSP3D’s discriminator-guided pseudo-labeling relies heavily on adversarial training, which can be unstable \cite{SSP3D}, while SSMP uses fixed weighting of fused priors without adapting to sample reliability \cite{SSMP}. Thus, a dynamic, geometry-aware weighting mechanism is needed to filter out unreliable SDF predictions and improve implicit reconstruction accuracy.

\section{Method}

\begin{figure}[t]
  \centering
  \includegraphics[width=\linewidth]{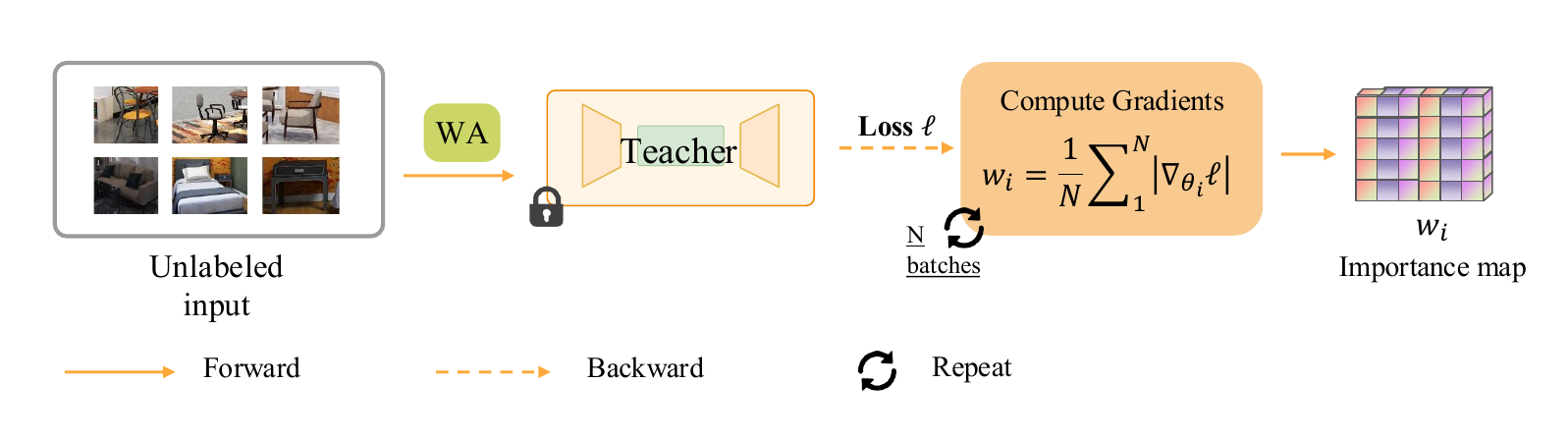}
  \caption{Gradient-guided importance map computation.}
  \label{fig:importance}
\end{figure}
\subsection{Adaptive Pseudo‐label Weighting Strategy}

In this section, we describe the proposed MetaSSP framework in detail.  Section~\ref{sec:overview} gives an overview of the two-stage pipeline.  Sections~\ref{sec:importance}--\ref{sec:ema} present the core components: gradient-based parameter importance, adaptive pseudo-label weighting, and meta-adaptive EMA with importance regularization.  Finally, Section~\ref{sec:data augmentation} introduce the strategy by using different image augmentation methods to achieve geometric and photometric consistency.

\subsection{Framework Overview}
\label{sec:overview} 
As shown in Figure~\ref{fig:framework}, our semi-supervised implicit 3D reconstruction framework consists of two stages. In the first stage, Warm-up, we perform full-supervision pre-training of the teacher network \(T_0\) using only 10\% of the available labeled data. Specifically, we adopt the SSR implicit SDF architecture and train for a fixed number of epochs on the labeled subset to obtain an initial teacher model capable of mapping RGB images to SDF \cite{ssr,marchingcube}. This warm-up provides reliable feature representations and initial SDF predictions for subsequent pseudo-label generation.

In the second stage, Stage2 Semi-supervised Training, we leverage abundant unlabeled data to continually refine the student network \(S\), while propagating its improvements back to the teacher via an EMA update. At the start of each epoch, we estimate per-parameter importance of the current teacher model on unlabeled samples. Next, for each unlabeled image, we perform both weak and strong augmentations, obtain two SDF outputs from the teacher, and combine their consistency loss with the variance of the weakly augmented SDF to compute an adaptive pseudo-label weight. The student is updated by minimizing a weighted sum of the supervised loss on labeled data and the pseudo-supervision loss on unlabeled data. Finally, we update the teacher model using a meta-adaptive EMA mechanism regularized by the parameter importance weights. This loop continues until convergence, at which point the best-performing teacher model is selected.

\subsection{Gradient‐based Parameter Importance Estimation}
\label{sec:importance} 
To safeguard the teacher model from catastrophic drift, we assess each parameter’s influence on the implicit SDF predictions over a small unlabeled subset every epoch (Fig.~\ref{fig:importance}).  Concretely, we draw $N=100$ batches of unlabeled images, apply the same weak augmentation used for pseudo‐labeling, and compute the signed‐distance output $\mathrm{sdf} = T(\mathrm{WA}(x))$.  We then form a temporary scalar loss  
\begin{equation}
\ell = \|\mathrm{sdf}\|_2^2 = \sum_{j=1}^M \mathrm{sdf}_j^2,
\end{equation}  
which—though not the final reconstruction objective—serves to backpropagate a gradient signal through every weight.  For each parameter $\theta_i$, we accumulate the absolute gradient magnitude $|\partial\ell/\partial\theta_i|$ across all $N$ batches and divide by $N$, yielding the importance weight $\omega_i$.  Intuitively, parameters with larger $\omega_i$ exert stronger control over the SDF field and are therefore protected during our meta‐adaptive EMA updates.

To generate reliable pseudo‐labels, we exploit both augmentation consistency and the intrinsic SDF distribution on unlabeled inputs.  Concretely, for each image \(x\) we first obtain two teacher‐network predictions under weak and strong augmentations:
\begin{equation}
\hat f_w = T\bigl(\mathrm{WA}(x)\bigr),\quad
\hat f_s = T\bigl(\mathrm{SA}(x)\bigr).
\end{equation}

We then compute a consistency measure by applying our usual pseudo‐loss function \(\mathcal{L}\):
\begin{equation}
\ell_{\mathrm{cons}} \;=\;\mathcal{L}\bigl(\hat f_s,\hat f_w\bigr),
\end{equation}

where a smaller \(\ell_{\mathrm{cons}}\) indicates that the pseudo‐labels are stable under heavy perturbation.  In parallel, we flatten \(\hat f_w\) into a vector and compute its sample variance,
since low variance implies the SDF values concentrate near the true surface.  To fuse these two criteria into a single quality score, we linearly combine and clamp into \([0,1]\):
\begin{equation}
w_{\mathrm{pseudo}}
\;=\;\operatorname{clip}\Bigl(1 \;-\;\alpha\,\ell_{\mathrm{cons}}
\;-\;\beta\,\sigma^2\,,\,0,1\Bigr),
\label{eq:pseudo_weight}
\end{equation}

 with hyperparameters \(\alpha=\beta=4\) (so that \(w_{\mathrm{pseudo}}\le0.4\)). Finally, during student training we blend the supervised loss \(L_{\mathrm{sup}}\) and the pseudo‐supervision loss \(L_{\mathrm{unsup}}\) for each loss term \(k\) via:
\begin{equation}
L
\;=\;\sum_{k}
\Bigl[(1 - \lambda\,w_{\mathrm{pseudo}})\,L_{\mathrm{sup},k}
\;+\;\lambda\,w_{\mathrm{pseudo}}\,L_{\mathrm{unsup},k}\Bigr],
\label{eq:final_loss}
\end{equation}

where \(\lambda=0.2\) controls the maximum contribution of pseudo‐labels.  In effect, Eq.~\eqref{eq:pseudo_weight} automatically down‐weights noisy pseudo‐labels (high inconsistency or high variance) and amplifies those that are both consistent and geometrically concentrated, fully leveraging the implicit SDF representation’s sensitivity to surface proximity.

\begin{table*}[t]
  \centering
  \scriptsize
  \setlength{\tabcolsep}{4pt}
  \caption{Comparisons of single-view 3D object reconstruction on Pix3D with 10\% supervised data. We report the mean IoU (\%) of all categories. The best number for each category is highlighted in bold.}
  \label{tab:pix3d_iou}
  \begin{tabularx}{\linewidth}{lYYYYYYYYYY}
    \toprule
    \toprule
    Approach / split & chair & bed & bookcase & desk & misc & sofa & table & tool & wardrobe & Mean \\
    & 267/2672 & 78/781 & 28/282 & 54/546 & 4/48 & 153/1532 & 145/1451 & 3/36 & 18/189 & \\
    \midrule
    \makecell{Supervised~\cite{ssr}}
      & 16.10 & 31.25 & 29.72 & 22.42 & 19.88 & 67.81 & 18.10 & 16.69 & 79.25 & 29.31 \\
    \midrule[0.75pt]
    \makecell{MeanTeacher~\cite{MeanTeacher}\\{\scriptsize\color{red}(+7.13\%)}}
      & \makecell{21.66\\{\color{red}\(\uparrow5.56\)}}
      & \makecell{35.04\\{\color{red}\(\uparrow3.79\)}}
      & \makecell{18.88\\{\color{blue}\(\downarrow10.84\)}}
      & \makecell{26.17\\{\color{red}\(\uparrow3.75\)}}
      & \makecell{\underline{22.37}\\{\color{red}\(\uparrow2.49\)}}
      & \makecell{64.19\\{\color{blue}\(\downarrow3.62\)}}
      & \makecell{\underline{24.03}\\{\color{red}\(\uparrow5.93\)}}
      & \makecell{ 9.18\\{\color{blue}\(\downarrow7.51\)}}
      & \makecell{84.34\\{\color{red}\(\uparrow5.09\)}}
      & \makecell{31.40\\{\color{red}\(\uparrow2.09\)}} \\
    \midrule[0.75pt]
    \makecell{FixMatch~\cite{FixMatch} \\{\scriptsize\color{red}(+3.55\%)}}
      & \makecell{21.95\\{\color{red}\(\uparrow5.85\)}}
      & \makecell{26.69\\{\color{blue}\(\downarrow4.56\)}}
      & \makecell{16.06\\{\color{blue}\(\downarrow13.66\)}}
      & \makecell{22.12\\{\color{blue}\(\downarrow0.30\)}}
      & \makecell{17.87\\{\color{blue}\(\downarrow2.01\)}}
      & \makecell{63.74\\{\color{blue}\(\downarrow4.07\)}}
      & \makecell{20.64\\{\color{red}\(\uparrow2.54\)}}
      & \makecell{ 6.89\\{\color{blue}\(\downarrow9.80\)}}
      & \makecell{84.45\\{\color{red}\(\uparrow5.20\)}}
      & \makecell{30.35\\{\color{red}\(\uparrow1.04\)}} \\
    \midrule[0.75pt]
    \makecell{SSP3D~\cite{SSP3D}\\{\scriptsize\color{red}(+20.75\%)}}
      & \makecell{\underline{23.97}\\{\color{red}\(\uparrow7.87\)}}
      & \makecell{\textbf{46.33}\\{\color{red}\(\uparrow15.08\)}}
      & \makecell{\underline{32.77}\\{\color{red}\(\uparrow3.05\)}}
      & \makecell{\textbf{32.89}\\{\color{red}\(\uparrow10.47\)}}
      & \makecell{\textbf{24.35}\\{\color{red}\(\uparrow4.47\)}}
      & \makecell{\underline{68.32}\\{\color{red}\(\uparrow0.51\)}}
      & \makecell{23.84\\{\color{red}\(\uparrow5.74\)}}
      & \makecell{\textbf{39.06}\\{\color{red}\(\uparrow22.37\)}}
      & \makecell{\textbf{89.59}\\{\color{red}\(\uparrow10.34\)}}
      & \makecell{\underline{35.39}\\{\color{red}\(\uparrow6.08\)}} \\
    \midrule[0.75pt]
    \makecell{Ours\\{\scriptsize\color{red}(+24.09\%)}}
      & \makecell{\textbf{25.35}\\{\color{red}\(\uparrow9.25\)}}
      & \makecell{\underline{40.05}\\{\color{red}\(\uparrow8.80\)}}
      & \makecell{\textbf{39.65}\\{\color{red}\(\uparrow9.93\)}}
      & \makecell{\underline{28.85}\\{\color{red}\(\uparrow6.43\)}}
      & \makecell{16.83\\{\color{blue}\(\downarrow3.05\)}}
      & \makecell{\textbf{69.63}\\{\color{red}\(\uparrow1.82\)}}
      & \makecell{\textbf{24.38}\\{\color{red}\(\uparrow6.28\)}}
      & \makecell{\underline{18.26}\\{\color{red}\(\uparrow1.57\)}}
      & \makecell{\underline{86.64}\\{\color{red}\(\uparrow7.39\)}}
      & \makecell{\textbf{36.37}\\{\color{red}\(\uparrow7.06\)}} \\
    \bottomrule
    \bottomrule
  \end{tabularx}
\end{table*}

\begin{figure}[t]
  \centering
  \includegraphics[width=\linewidth]{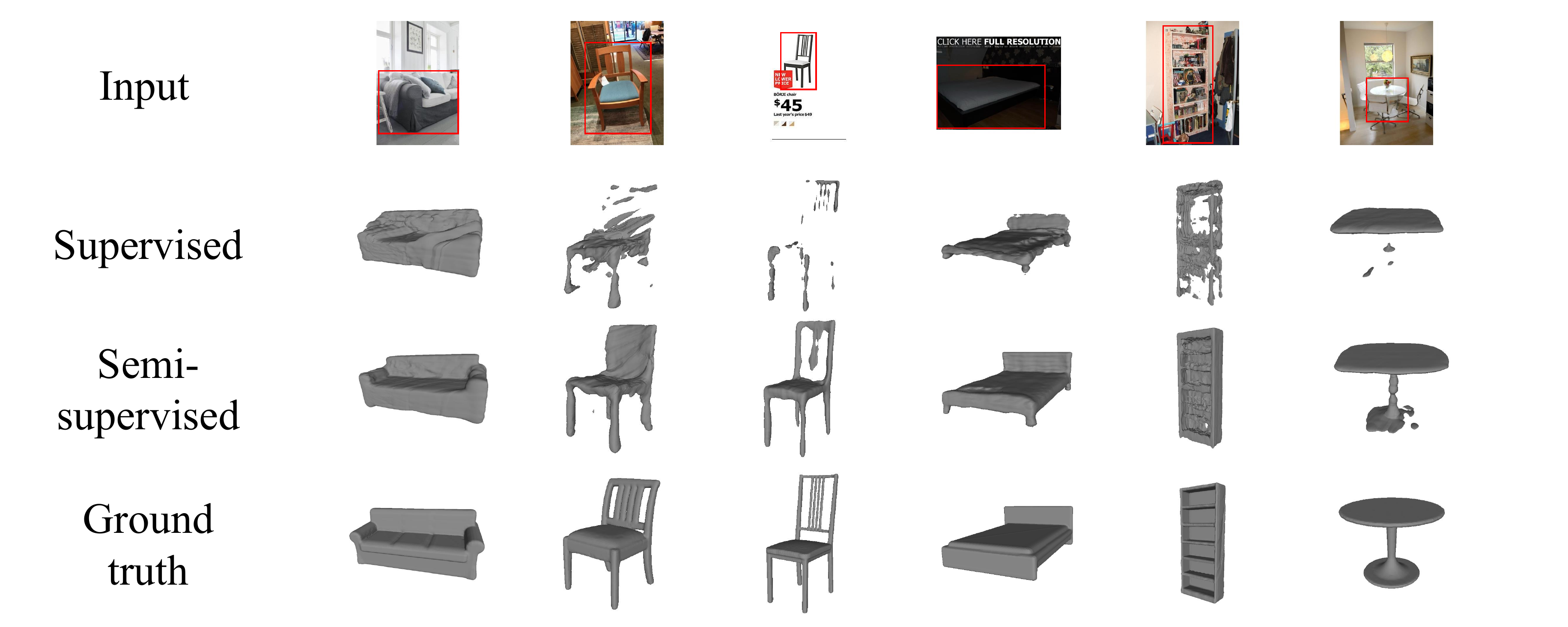}
  \caption{Results come from warm-up supervised, semi- supervised and ground truth.}
  \label{fig:output}
\end{figure}

\subsection{Meta‐adaptive EMA with Parameter Importance Regularization}
\label{sec:ema} 

Fixed‐momentum EMA updates:
\begin{equation}
\theta_T^{(t+1)} \;=\; m\,\theta_T^{(t)} + (1-m)\,\theta_S^{(t)},
\end{equation}

it often fails when the student and teacher parameters diverge, leading to teacher degradation.  To adapt automatically, we first compute a base momentum via cosine annealing over the current step \(t\) in a total of \(T\) steps:
\begin{equation}
m_{\mathrm{base}}
=1 - (1 - m_0)\,\frac{\cos\!\bigl(\pi\,t/T\bigr) + 1}{2},
\end{equation}

where \(m_0\) is the initial EMA momentum.  We then feed the teacher–student SDF loss gap \(\Delta\ell = \ell_T - \ell_S\), the teacher’s evaluation loss \(\ell_T\), and the normalized epoch index \(t/T\) into a small MLP meta‐controller, whose output is
\begin{equation}
\gamma = \sigma\bigl(W_2\!\bigl(\mathrm{ReLU}(W_1[\Delta\ell,\ell_T,t/T])\bigr)\bigr)\times0.01 + 0.995,
\end{equation}

so that \(\gamma\) lies in \([0.96,\,0.9996]\).  The actual EMA momentum is then clamped as
\begin{equation}
m = \mathrm{clip}\bigl(\gamma\,m_{\mathrm{base}},\,m_{\min},\,m_{\max}\bigr).
\end{equation}

Next, to prevent catastrophic forgetting of critical parameters, we incorporate the importance weights \(\{\omega_i\}\) from Section~\ref{sec:importance} directly into the update.  Denoting the parameter‐wise difference \(\Delta\theta_i=\theta_{S,i}-\theta_{T,i}\), we regularize each update by
\begin{equation}
\theta_{T,i}^{(t+1)}
=\theta_{T,i}^{(t)} \;+\;\frac{1-m}{1+\eta\,\omega_i}\,\Delta\theta_i,
\label{eq:reg_ema}
\end{equation}
where \(\eta>0\) is a small hyperparameter.  In effect, parameters with large \(\omega_i\) (high importance) receive a smaller fraction of the student update, preserving essential SDF reconstruction capability.

Finally, to guard against cases where the student and teacher have drifted too far—evidenced by \(\ell_{\mathrm{sup}} - \ell_T < \delta\) on the validation set—we occasionally apply a “reset” update with a lower effective momentum (e.g.\ \(0.6\,m\)), shrinking the gap and restoring EMA gains.  This two‐fold adaptation—dynamic momentum scheduling and importance‐regularized updates—ensures the teacher remains both stable and responsive throughout semi‐supervised training.

\begin{table}[ht]
  \centering
    \caption{Hyperparameter settings for the warm‐up and semi‐supervised phases.}
  \resizebox{\columnwidth}{!}{
  \begin{tabular}{lcc}
    \toprule
    Parameter                 & Warm‐up Phase                & Semi‐supervised Phase \\
    \midrule
    Data split                & 10\% labeled Pix3D           & 10\% labeled + 90\% unlabeled Pix3D \\
    Number of epochs          & 400                           & 200                     \\
    Effective batch size      & 128                           & 160                     \\
    Scheduler decay epochs    & [350, 390]                      & [170, 190]                \\
    \bottomrule
  \end{tabular}
  }

  \label{tab:training_settings}
\end{table}

\subsection{Data Augmentation for Consistency Regularization}
\label{sec:data augmentation} 
To enforce geometric and photometric consistency in our semi-supervised framework, we apply two levels of augmentation to each unlabeled image:

  \textbf{Weak Augmentation (WA):} mild color‐jitter (\(\pm10\%\) brightness/contrast), small rotations (\(\pm20^\circ\)) and light gaussian noise.  WA ensures that the teacher’s baseline predictions do not drift when presented with slight perturbations.

  \textbf{Strong Augmentation (SA):} aggressive color shifts (up to \(\pm40\%\)), random Gaussian blur, occlusion patches (1–4 random erasing blocks), plus white‐noise and hue shifts.  SA challenges the student to produce stable pseudo-labels under heavy distortion.

We denote the two transformed versions of \(x\) by \(\mathrm{WA}(x)\) and \(\mathrm{SA}(x)\).  In Section~\ref{sec:ema} these are used to compute our consistency loss \(\ell_{\mathrm{cons}}\) and drive the adaptive pseudo‐label weighting.

\begin{table}[ht]
  \centering
    \caption{Ablation configurations and their components.}
  \label{tab:ablation_settings}
  \resizebox{\columnwidth}{!}{%
  \begin{tabular}{lccc}
    \toprule
    Configuration     & EMA                & Importance Regularization & Pseudo-label Weighting \\
    \midrule
    Baseline          & No                 & No                        & No            \\
    EMA-fixed-1         & Yes (0.996)& No                        & Fixed (0.5)            \\
    EMA-fixed-2         & Yes (0.996)& No                        & Fixed (0.2)            \\
    ImpEMA-fixed         & Yes (0.996)& Yes                        & Fixed (0.2)            \\
    Dyn-ImpEMA-fixed      & Yes (dynamic)      & Yes                       & Fixed (0.2)            \\
    Dyn-ImpEMA-adaptive   & Yes (dynamic)      & Yes                       & Adaptive               \\
    \bottomrule
  \end{tabular}%
  }
\end{table}

\begin{table*}[t]
  \centering
  \caption{3D reconstruction on the Pix3D dataset using 10\% of the data, based on L1 CD, with the results multiplied by $10^2$.}
  \label{tab:pix3d_chamfer}
  \resizebox{\linewidth}{!}{
  \begin{tabular}{lccccccccc}
    \toprule
    Method & bed & bookcase & desk & misc & sofa & table & tool & wardrobe & Average \\
    \midrule
    3D attributeflow~\cite{3DAttriFlow} 
      & 9.19 & 7.99 & 7.16 & 13.60 & 5.68 & 8.61 & 14.40 & 5.16 & 7.59 \\
    Pixel2Point~\cite{Pixel2point} 
      & 11.16 & 8.04 & 8.49 & 12.20 & 6.42 & 10.15 & 12.18 & 6.69 & 8.78 \\
    Pix2Voxel~\cite{Pix2Vox} 
      & 9.47 & 7.43 & 7.87 & \underline{10.21} & 6.37 & 8.49 & 12.39 & 6.04 & 8.53 \\
    SSMP~\cite{SSMP} 
      & \underline{6.73} & \underline{6.48} & \underline{7.11} & 13.44 & \underline{4.60} & \underline{8.07} & \underline{11.66} & \underline{3.42} & \underline{6.53} \\
    \textbf{Ours }
      & \textbf{1.72} & \textbf{0.61} & \textbf{2.59 }& \textbf{8.32} & \textbf{0.58} & \textbf{5.74} & \textbf{2.47} & \textbf{0.14} & \textbf{3.12} \\

    \bottomrule
  \end{tabular}
  }
\end{table*}

\section{Experiments}
In this section, we first introduce the dataset and evaluation metrics (Section~\ref{sec:data}), then detail our implementation and training setup (Section~\ref{sec:implement}). We proceed with ablation studies to validate the contributions of each component (Section~\ref{sec:ablation}).

\subsection{Dataset and Evaluation Metrics}
\label{sec:data} 
All experiments are conducted on the Pix3D dataset~\cite{pix3d}, which comprises real‐world single‐view RGB images precisely aligned with 3D CAD models. We follow the standard “S1” split of Pix3D~\cite{pix3d}, using 7\,539 images for training and 2\,530 images for validation and testing (the validation and test sets are identical)~\cite{pix3d}. Reconstruction quality is primarily measured by CD, which computes the average bidirectional point‐to‐surface distance, and Intersection over Union (IoU), which evaluates volumetric overlap between predicted and ground‐truth meshes. To quantify the gains from the Warm‐up to the semi‐supervised stages, we also report the F-score (at a threshold of 1\,\% of the bounding‐box diagonal), NC(the average cosine similarity between predicted and ground‐truth surface normals), and PSNR on rendered depth maps to assess fine‐detail fidelity.

\subsection{Implementation Details}
\label{sec:implement} 
All experiments were conducted on a Pegasus with NVIDIA H100 GPU, running Python 3.10.15, PyTorch v2.4.1+cu124, CUDA 12.4 and cuDNN 9.1.0. Training is divided into two successive phases: a fully supervised “warm‐up” and a semi‐supervised joint stage. Table \ref{tab:training_settings} summarizes the key hyperparameters for each phase.

\subsection{Ablation Studies}
\label{sec:ablation} 
To disentangle the contributions of each component in MetaSSP, we designed five ablation configurations (Table~\ref{tab:ablation_settings}). Starting from a pure warm-up baseline without any teacher smoothing or pseudo-label weighting, we first enabled a fixed‐momentum EMA (0.996) while keeping the pseudo‐label weight constant at 0.5 (EMA-fixed-1) or 0.2 (EMA-fixed-2). We then introduced our gradient‐guided importance regularization (ImpEMA-fixed), followed by a dynamic EMA schedule with fixed weighting (Dyn-ImpEMA-fixed), and finally combined dynamic EMA with our adaptive pseudo‐label weighting (Dyn-ImpEMA-adaptive).

\section{Results and Analysis}
\subsection{Comparison Between Warm-up and Semi-Supervised Fine-tuning}

Table~\ref{tab:Pix3D} together with Figure~\ref{fig:output} compares the reconstruction quality after the supervised warm-up stage versus our MetaSSP semi-supervised fine-tuning on 10\% labeled Pix3D. MetaSSP yields clear improvements in both global shape completeness and local detail fidelity: CD drops from 3.93 to 3.12, and IoU rises from 29.31\% to 33.11\%. These gains demonstrate that our framework effectively leverages unlabeled data under limited supervision and that gradient-guided importance regularization together with SDF-aware pseudo-label weighting substantially enhances shape recovery. However, the improvements are not uniform: the “misc” category, with extremely few labels, shows degraded performance, indicating that residual pseudo-label noise persists in the most underrepresented classes. In future work, we will refine confidence thresholds and design targeted augmentations to close this remaining gap.

\begin{table*}[t]
  \centering
  \caption{The evaluation results between supervised warm-up training and semi-supervised training.}
  \label{tab:Pix3D}
  \resizebox{\linewidth}{!}{ 
    \begin{tabular}{@{}llcccccccccc@{}}
      \toprule[1pt]
      \toprule[1pt]
      Metrics & Models & bed & bookcase & chair & desk & sofa & table & tool & wardrobe & misc & mean \\
      \midrule[0.75pt]
      \multirow{2}{*}{\shortstack{CD $\downarrow$ \\ \textcolor{red}{(20.61\%)}}} 
      & Supervised  & 2.21 & 1.02 & 5.28 & 3.16 & 0.69 & 6.64 & \textbf{1.65} & 0.37 & \textbf{5.52} &  3.93\\
      & SemiSup& \textbf{1.72} & \textbf{0.61} & \textbf{3.85} & \textbf{2.59} & \textbf{0.58} & \textbf{5.74} & 2.47 & \textbf{0.14} & 8.32 & \textbf{3.12} \\
      \midrule[0.75pt]
      \multirow{2}{*}{\shortstack{F-Score $\uparrow$ \\ \textcolor{red}{(15.6\%)}}} 
      & Supervised  &  44.55 & 58.72 & 34.68 & 49.13 & 66.72 & 44.43 & \textbf{52.78} & 85.88 & \textbf{27.43} & 46.69\\
      & SemiSup & \textbf{49.74} & \textbf{69.31} & \textbf{45.50} & \textbf{54.16} & \textbf{69.99} & \textbf{50.33} & 47.00 & \textbf{94.98} & 26.55 & \textbf{53.95}\\
      \midrule[0.75pt]
      \multirow{2}{*}{\shortstack{NC $\uparrow$ \\ \textcolor{red}{(6.4\%)}}} 
      & Supervised & 0.742 & 0.670 & 0.636 & 0.761 & 0.853 & 0.812 & 0.593 & 0.935 & 0.580 & 0.737\\
      & SemiSup & \textbf{0.791} & \textbf{0.752} & \textbf{0.702} & \textbf{0.791} & \textbf{0.890} & \textbf{0.833} & \textbf{0.599} & \textbf{0.978} & \textbf{0.603} & \textbf{0.784}\\
      \midrule[0.75pt]
      \multirow{2}{*}{\shortstack{PSNR $\uparrow$ \\ \textcolor{red}{(4.8\%)}}} 
      & Supervised & 23.48 & 27.14 & 21.25 & 22.13 & 26.73 & 21.81 & \textbf{24.57} & 29.09 & \textbf{17.93} & 23.06\\
      & SemiSup & \textbf{24.08} & \textbf{27.23} & \textbf{22.65} & \textbf{22.88} & \textbf{27.27} & \textbf{23.33} & 23.21 & \textbf{32.60} & 17.53 & \textbf{24.17} \\
      \bottomrule[1pt]
      \bottomrule[1pt]
    \end{tabular}
  }
\end{table*}

\subsection{Comparison with Existing Methods}

Table~\ref{tab:pix3d_iou} reports per-category mean IoU and Table~\ref{tab:pix3d_chamfer} shows L1 CD (×$10^2$) on Pix3D with 10\% labels, for MetaSSP and four prominent baselines. MetaSSP achieves 36.37\% IoU and 3.12 Chamfer, outperforming MeanTeacher \cite{MeanTeacher}, FixMatch \cite{FixMatch}, SSP3D \cite{SSP3D}, and fully-supervised approaches \cite{ssr}. By contrast, MeanTeacher’s uniform EMA averaging tends to accumulate low-quality pseudo-labels, FixMatch’s fixed confidence threshold cannot adapt to geometric variance in complex shapes, and SSP3D \cite{SSP3D} lacks gradient-guided filtering of noisy samples. In MetaSSP, dynamic EMA combined with gradient-guided importance regularization stabilizes teacher–student updates around high-confidence regions, while SDF-aware weighting fuses variance information to suppress noisy labels and sharpen supervision. This synergy enables faithful reconstruction of both global structure and fine details under scarce labels.

\subsection{Ablation Study}

To isolate each component’s effect, Table~\ref{tab:ablation_settings} defines six ablation configurations and Table~\ref{tab:ablation} reports their performance after only 50 epochs of semi-supervised fine-tuning. Applying a fixed-momentum EMA alone (without importance regularization or adaptive weighting) actually degrades performance—Chamfer increases and IoU falls—because it indiscriminately aggregates noisy teacher predictions. Introducing gradient-guided importance regularization recovers and surpasses the warm-up baseline. Finally, combining dynamic EMA with adaptive pseudo-label weighting (Dyn-ImpEMA-adaptive) delivers the best results (IoU 33.11\%, Chamfer 3.26), confirming that a continually refined teacher model and flexible weighting strategy are essential for robust gains under limited labels.

\begin{table}[ht]
  \centering
  \caption{Ablation study on Pix3D with 50 epochs.}
  \label{tab:ablation}
  \resizebox{\linewidth}{!}{
  \begin{tabular}{lccccc}
    \toprule
    Method                   & CD (\(\downarrow\)) & Fscore (\(\uparrow\)) & psnr (\(\uparrow\)) & IoU (\(\uparrow\)) & NC (\(\uparrow\)) \\
    \midrule
    Baseline       & 3.93    & 46.69      & 23.06     & 29.31   & 0.737        \\
    EMA-fixed-1      & 6.24    & 21.73    & 15.12      & 23.00  & 0.544       \\
    EMA-fixed-2      & 4.51    & 33.72    & 19.40      & 25.44   & 0.698       \\
    ImpEMA-fixed   & 3.53       & 48.77    & 23.49     & 32.16      & 0.755      \\
    Dyn-ImpEMA-fixed   & 3.37      & 50.13    & 23.63   &  32.87     & 0.752      \\
    \textbf{Dyn-ImpEMA-adaptive} & \textbf{3.26}    & \textbf{51.83}  & \textbf{24.01}  & \textbf{33.11} & \textbf{0.767}    \\
    \bottomrule
  \end{tabular}
  }
\end{table}

Table~\ref{tab:ablation} shows that each component—EMA stabilization via importance regularization and adaptive pseudo-label weighting—contributes to consistent improvements in both CD and IoU.

\section{Conclusion and Future Work}
We presented MetaSSP, a semi-supervised framework for SDF-based single-view 3D reconstruction that effectively bridges the gap between supervised and unlabeled data. By integrating gradient-guided importance regularization with a meta-adaptive EMA controller, MetaSSP stabilizes teacher–student learning and ensures consistent geometric supervision. Complemented by an SDF-aware pseudo-label weighting strategy, our framework achieves state-of-the-art performance on the Pix3D benchmark across multiple metrics, demonstrating strong robustness under limited supervision. 

In future work, we aim to extend MetaSSP to diverse label ratios and incorporate category-aware SDF priors to further enhance its scalability and generalization to complex real-world scenes.

\textbf{Acknowledgment.}~This work was supported by MEXT Promotion of Development of a Joint Usage/Research System Project: Coalition of Universities for Research Excellence Program (CURE) Grant Number JPMXP1323015474.

{ 
    \small
    \bibliographystyle{IEEEtran}
    \bibliography{main}
}

\end{document}